\PassOptionsToPackage{hyphens}{url}
\documentclass[11pt,letterpaper]{article}
\usepackage[nohyperref]{emnlp2017}
\usepackage{times}
\usepackage{latexsym}
\usepackage{amsmath}
\usepackage{amssymb}
\usepackage{graphicx}
\usepackage{placeins}
\usepackage{color}
\usepackage{hyperref}
\usepackage{url}

\emnlpfinalcopy

\def\emnlppaperid{***}

\title{Explaining Recurrent Neural Network Predictions in Sentiment Analysis}

\author{Leila Arras$^1$, Gr{\'e}goire Montavon$^2$, Klaus-Robert M{\"u}ller$^{2,3,4}$, and Wojciech Samek$^1$\\
  $^1$Machine Learning Group, Fraunhofer Heinrich Hertz Institute, Berlin, Germany \\
  $^2$Machine Learning Group, Technische Universit\"at Berlin, Berlin, Germany \\
  $^3$Department of Brain and Cognitive Engineering, Korea University, Seoul, Korea \\
  $^4$Max Planck Institute for Informatics, Saarbr{\"u}cken, Germany\\
  {\tt \{leila.arras, wojciech.samek\}@hhi.fraunhofer.de}}

\date{}

\begin{document}

\maketitle

\begin{abstract}
Recently, a technique called Layer-wise Relevance Propagation (LRP) was shown to deliver insightful {\it explanations} in the form of input space relevances for understanding feed-forward neural network classification decisions.
In the present work, we extend the usage of LRP to recurrent neural networks. We propose a specific propagation rule applicable to multiplicative connections as they arise in recurrent network architectures such as LSTMs and GRUs.
We apply our technique to a word-based bi-directional LSTM model on a five-class sentiment prediction task, and evaluate the resulting LRP relevances both qualitatively and quantitatively, obtaining better results than a gradient-based related method which was used in previous work.
\end{abstract}

\section{Introduction}

Semantic composition plays an important role in sentiment analysis of phrases and sentences. This includes detecting the scope and impact of negation in reversing a sentiment's polarity, as well as quantifying the influence of modifiers, such as degree adverbs and intensifiers, in rescaling the sentiment's intensity \cite{Challenges_SA}.

Recently, a trend emerged for tackling these challenges via deep learning models such as convolutional and recurrent neural networks, as observed e.g. on the SemEval-2016 Task for {\it Sentiment Analysis in Twitter} \cite{SemEval_2016}.

As these models become increasingly predictive, one also needs to make sure that they work as intended, in particular, their decisions should be made as transparent as possible.

Some forms of transparency are readily obtained from the structure of the model, e.g. recursive nets \cite{Socher-etal:2013}, where sentiment can be probed at each node of a parsing tree. 

Another type of analysis seeks to determine what input features were important for reaching the final top-layer prediction.
Recent work in this direction has focused on bringing measures of feature importance to state-of-the-art models such as deep convolutional neural networks for vision \cite{Simonyan, DBLP:conf/eccv/ZeilerF14, Bach_LRP, DBLP:conf/kdd/Ribeiro0G16}, or to general deep neural networks for text \cite{Denil,Li_2016a,arras-acl16,Li_2016b,MurdochS17}.

Some of these techniques are based on the model's local gradient information while other methods seek to redistribute the function's value on the input variables, typically by reverse propagation in the neural network graph \cite{DBLP:conf/cidm/LandeckerTBMKB13, Bach_LRP, Montavon_2017}.
We refer the reader to \cite{MonArXiv17} for an overview on methods for understanding and interpreting deep neural network predictions.

\citet{Bach_LRP} proposed specific propagation rules for neural networks (LRP rules). These rules were shown to produce better explanations than e.g.\ gradient-based techniques \citep{Samek_TNNLS}, and were also successfully transferred to neural networks for text data \cite{Arras_2016b}.

In this paper, we extend LRP with a rule that handles multiplicative interactions in the LSTM model, a particularly suitable model for modeling long-range interactions in texts such as those occurring in sentiment analysis.

We then apply the extended LRP method to a bi-directional LSTM trained on a five-class sentiment prediction task. It allows us to produce reliable explanations of which words are responsible for attributing sentiment in individual texts,
compared to the explanations obtained by using a gradient-based approach.

\section{Methods}

Given a trained neural network that models a scalar-valued prediction function $f_c$ (also commonly referred to as a prediction score) for each class $c$ of a classification problem, and given an input vector $\boldsymbol{x}$, we are interested in computing for each input dimension $d$ of $\boldsymbol{x}$ a relevance score $R_d$ quantifying the relevance of $x_d$ w.r.t to a considered {\it target} class of interest $c$.
In others words, we want to analyze which features of $\boldsymbol x$ are important for the classifier's decision {\it toward} or {\it against} a class $c$.

In order to estimate the relevance of a {\it pool} of input space dimensions or variables (e.g. in NLP, when using distributed word embeddings as input, we would like to compute the relevance of a word, and not just of its single vector dimensions), we simply sum up the relevance scores $R_d$ of its constituting dimensions $d$.

In this described framework, there are two alternative methods to obtain the single input variable's relevance in the first place, which we detail in the following subsections.

\subsection{Gradient-based Sensitivity Analysis (SA)}
The relevances can be obtained by computing squared partial derivatives:
$$
R_d = \Big(\frac{\partial {f_c}}{\partial x_d}(\boldsymbol{x}) \Big)^2.
$$
For a neural network classifier, these derivatives can be obtained by standard gradient backpropagation \cite{rumelhart86}, and are made available by most neural network toolboxes. We refer to the above definition of relevance as Sensitivity Analysis (SA) \cite{Dimopoulos95,Gevrey03}. A similar technique was previously used in computer vision by \cite{Simonyan}, and in NLP by \cite{Li_2016a}.

Note that if we sum up the relevances of all input space dimensions $d$, we obtain the quantity $\|{\nabla}_{\boldsymbol x} \; f_c({\boldsymbol x})\|{_2^2}$, thus SA can be interpreted as a decomposition of the squared gradient norm.

\subsection{Layer-wise Relevance Propagation (LRP)}
Another technique to compute relevances was proposed in \cite{Bach_LRP} as the Layer-wise Relevance Propagation (LRP) algorithm. It is based on a layer-wise relevance conservation principle, and, for a given input $\boldsymbol{x}$, it redistributes the quantity $f_c(\boldsymbol{x})$, starting from the output layer of the network and backpropagating this quantity up to the input layer.
The LRP relevance propagation procedure can be described layer-by-layer for each type of layer occurring in a deep convolutional neural network (weighted linear connections following non-linear activation, pooling, normalization),
and consists in defining rules for attributing relevance to lower-layer neurons given the relevances of upper-layer neurons. Hereby each intermediate layer neuron gets attributed a relevance score, up to the input layer neurons.

In the case of recurrent neural network architectures such as LSTMs \cite{Hochreiter_1997} and GRUs \cite{Cho_2014}, there are two types of neural connections involved: many-to-one weighted linear connections, and two-to-one multiplicative interactions. Hence, we restrict our definition of the LRP procedure to these types of connections.
Note that, for simplification, we refrain from explicitly introducing a notation for non-linear activation functions; if such an activation is present at a neuron, we always take into account the {\it activated} lower-layer neuron's value in the subsequent formulas. 

In order to compute the input space relevances, we start by setting the relevance of the output layer neuron corresponding to the target class of interest $c$ to the value $f_c(\boldsymbol{x})$, and simply ignore the other output layer neurons (or equivalently set their relevance to zero). Then, we compute layer-by-layer a relevance score for each intermediate lower-layer neuron accordingly to one of the subsequent formulas, depending on the type of connection involved.

\noindent{\bf Weighted Connections.}
Let $z_j$ be an upper-layer neuron, whose value in the forward pass is computed as $z_j = \sum_{i}z_i \cdot w_{ij} + b_j$, where $z_i$ are the lower-layer neurons, and $w_{ij}$, $b_j$ are the connection weights and biases.

Given the relevances $R_j$ of the upper-layer neurons $z_j$, the goal is to compute the lower-layer relevances $R_i$ of the neurons $z_i$. (In the particular case of the output layer, we have a single upper-layer neuron $z_j$, whose relevance is set to its value, more precisely we set $R_j=f_c(\boldsymbol{x})$ to start the LRP procedure.) 
The relevance redistribution onto lower-layer neurons is performed in two steps. First, by computing relevance messages $R_{i \leftarrow j}$ going from upper-layer neurons $z_j$ to lower-layer neurons $z_i$. Then, by summing up incoming messages for each lower-layer neuron $z_i$ to obtain the relevance $R_i$.
The messages $R_{i \leftarrow j}$ are computed as a fraction of the relevance $R_j$ accordingly to the following rule:
$$
R_{i \leftarrow j} = \frac{z_i \cdot w_{ij}  + \frac{\epsilon \cdot {\text sign}(z_j) \; + \; \delta \cdot b_j}{N}}{z_j + \epsilon \cdot {\text sign} (z_j)} \; \cdot R_j
$$
where $N$ is the total number of lower-layer neurons to which $z_j$ is connected, $\epsilon$ is a small positive number which serves as a stabilizer (we use $\epsilon=0.001$ in our experiments), and ${\text sign}(z_j)=(1_{z_j \geq 0} - 1_{z_j < 0})$ is defined as the sign of $z_j$.
The relevance $R_i$ is subsequently computed as $R_i = \sum_{j} R_{i \leftarrow j}$.
Moreover, $\delta$ is a multiplicative factor that is either set to 1.0, in which case the total relevance of all neurons in the same layer is conserved, or else it is set to 0.0, which implies that a part of the total relevance is ``absorbed'' by the biases and that the relevance propagation rule is approximately conservative.
Per default we use the last variant with $\delta=0.0$ when we refer to LRP, and denote explicitly by LRP$_{cons}$ our results when we use $\delta=1.0$ in our experiments.

\noindent{\bf Multiplicative Interactions.}
Another type of connection is a two-way multiplicative interaction between lower-layer neurons.
Let $z_j$ be an upper-layer neuron, whose value in the forward pass is computed as the multiplication of the two lower-layer neuron values $z_g$ and $z_s$, i.e. $z_j = z_g \cdot z_s$.
In such multiplicative interactions, as they occur e.g. in LSTMs and GRUs, there is always one of the two lower-layer neurons that constitutes a {\it gate}, and whose value ranges between $[0,1]$ as the output of a sigmoid activation function (or in the particular case of GRUs, can also be equal to one minus a sigmoid activated value), we call it the $gate$ neuron $z_g$, and refer to the remaining one as the $source$ neuron $z_s$.

Given such a configuration, and denoting by $R_j$ the relevance of the upper-layer neuron $z_j$, we propose to redistribute the relevance onto lower-layer neurons in the following way: we set $R_g=0$ and $R_s=R_j$. 
The intuition behind this reallocation rule, is that the {\it gate} neuron decides already in the forward pass how much of the information contained in the {\it source} neuron should be retained to make the overall classification decision. Thereby the value $z_g$ controls how much relevance will be attributed to $z_j$ from upper-layer neurons. Thus, even if our local propagation rule seems to ignore the respective values of $z_g$ and $z_s$ to redistribute the relevance, these are indeed taken into account when computing the value $R_j$ from the relevances of the {\it next} upper-layer neurons to which $z_j$ is connected via weighted connections.   

\section{Recurrent Model and Data}
As a recurrent neural network model we employ a one hidden-layer bi-directional LSTM (bi-LSTM), trained on five-class sentiment prediction of phrases and sentences 
on the Stanford Sentiment Treebank movie reviews dataset \cite{Socher-etal:2013}, as was used in previous work on neural network interpretability \cite{Li_2016a} and made available by the authors\footnote{\url{https://github.com/jiweil/Visualizing-and-Understanding-Neural-Models-in-NLP}}.
This model takes as input a sequence of words $x_1, x_2,..., x_T$ (as well as this sequence in reversed order), where each word is represented by a word embedding of dimension 60, and has a hidden layer size of 60. A thorough model description can be found in the Appendix, and for details on the training we refer to \cite{Li_2016a}.

In our experiments, we use as input the 2210 tokenized sentences of the Stanford Sentiment Treebank test set \cite{Socher-etal:2013}, preprocessing them by lowercasing as was done in \cite{Li_2016a}.
On five-class sentiment prediction of full sentences (very negative, negative, neutral, positive, very positive) the model achieves 46.3\% accuracy, and for binary classification (positive vs. negative, ignoring neutral sentences) the test accuracy is 82.9\%.

Using this trained bi-LSTM, we compare two relevance decomposition methods: sensitivity analysis (SA) and Layer-wise Relevance Propagation (LRP). 
The former is similar to the ``First-Derivative Saliency'' used in \cite{Li_2016a}, besides that in their work the authors do not aggregate the relevance of single input variables to obtain a word-level relevance value (i.e. they only visualize relevance distributed over word embedding dimensions); moreover they employ the absolute value of partial derivatives (instead of squared partial derivatives as we do) to compute the relevance of single input variables.

In order to enable reproducibility and for encouraging further research, we make our implementation of both relevance decomposition methods available\footnote{\url{https://github.com/ArrasL/LRP_for_LSTM}} (see also \cite{LapJMLR16}).

\section{Results}

In this Section, we present qualitative as well as quantitative results we obtained by performing SA and LRP on test set sentences. 
As an outcome of the relevance decomposition for a chosen {\it target} class, we first get for each word embedding $x_t$ in an input sentence, a {\it vector} of relevance values. In order to obtain a {\it scalar} word-level relevance, we remind that we simply sum up the relevances contained in that vector.
Also note that, per definition, the SA relevances are positive while LRP relevances are signed.

\subsection{Decomposing Sentiment onto Words}

\begin{figure*}
\centering
\includegraphics[clip=true, trim=33mm 17.0cm 20mm 2.5cm, width=0.95\textwidth, resolution=300]{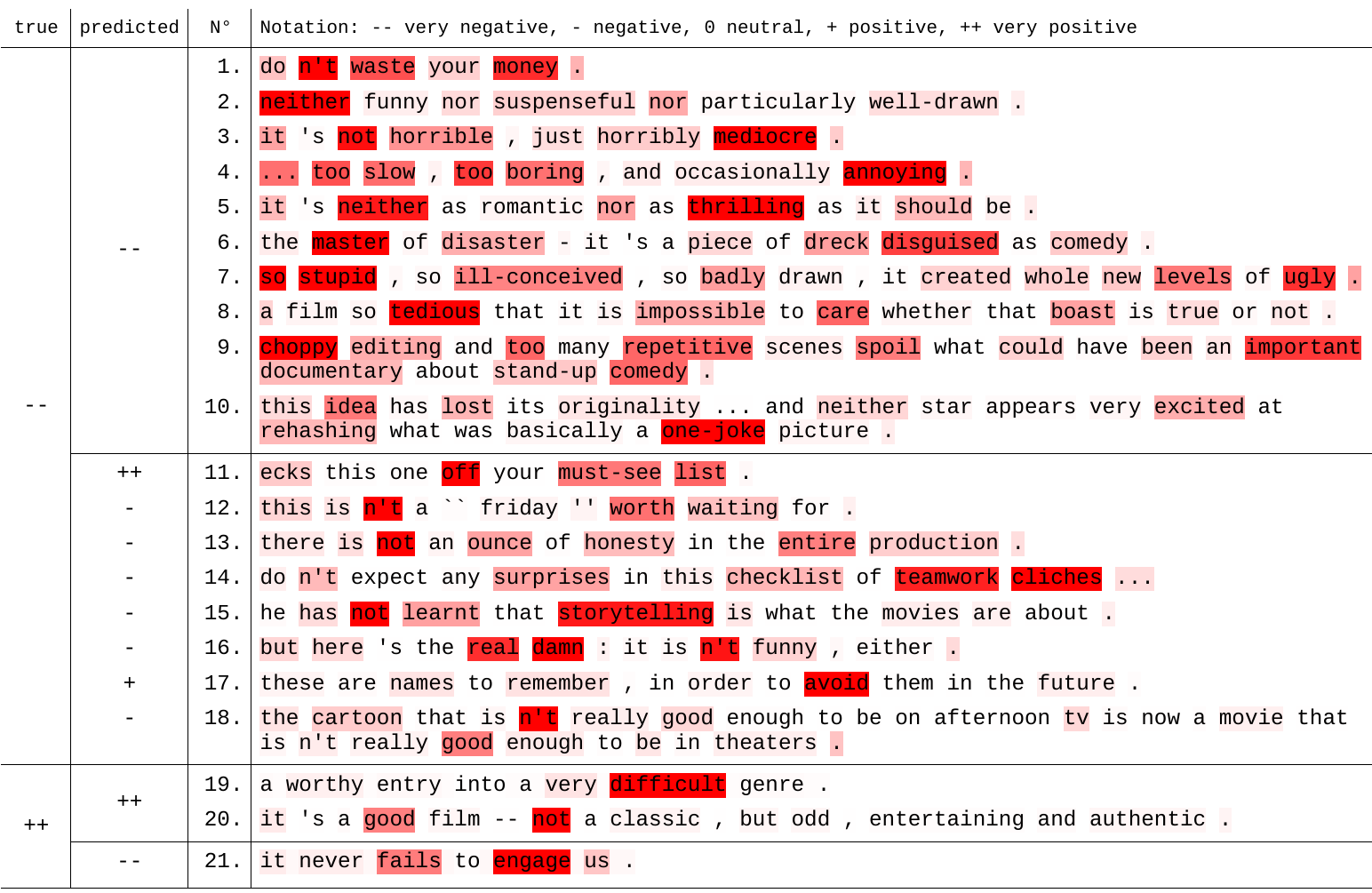}
\caption{ SA heatmaps of exemplary test sentences, using as target class the {\it true} sentence class. All relevances are positive and mapped to red, the color intensity is normalized to the maximum relevance per sentence. The true sentence class, and the classifier's predicted class, are indicated on the left.}\label{fig:heatmap_SA}
\end{figure*}

\begin{figure*}
\centering
\includegraphics[clip=true, trim=33mm 17.0cm 20mm 2.5cm, width=0.95\textwidth, resolution=300]{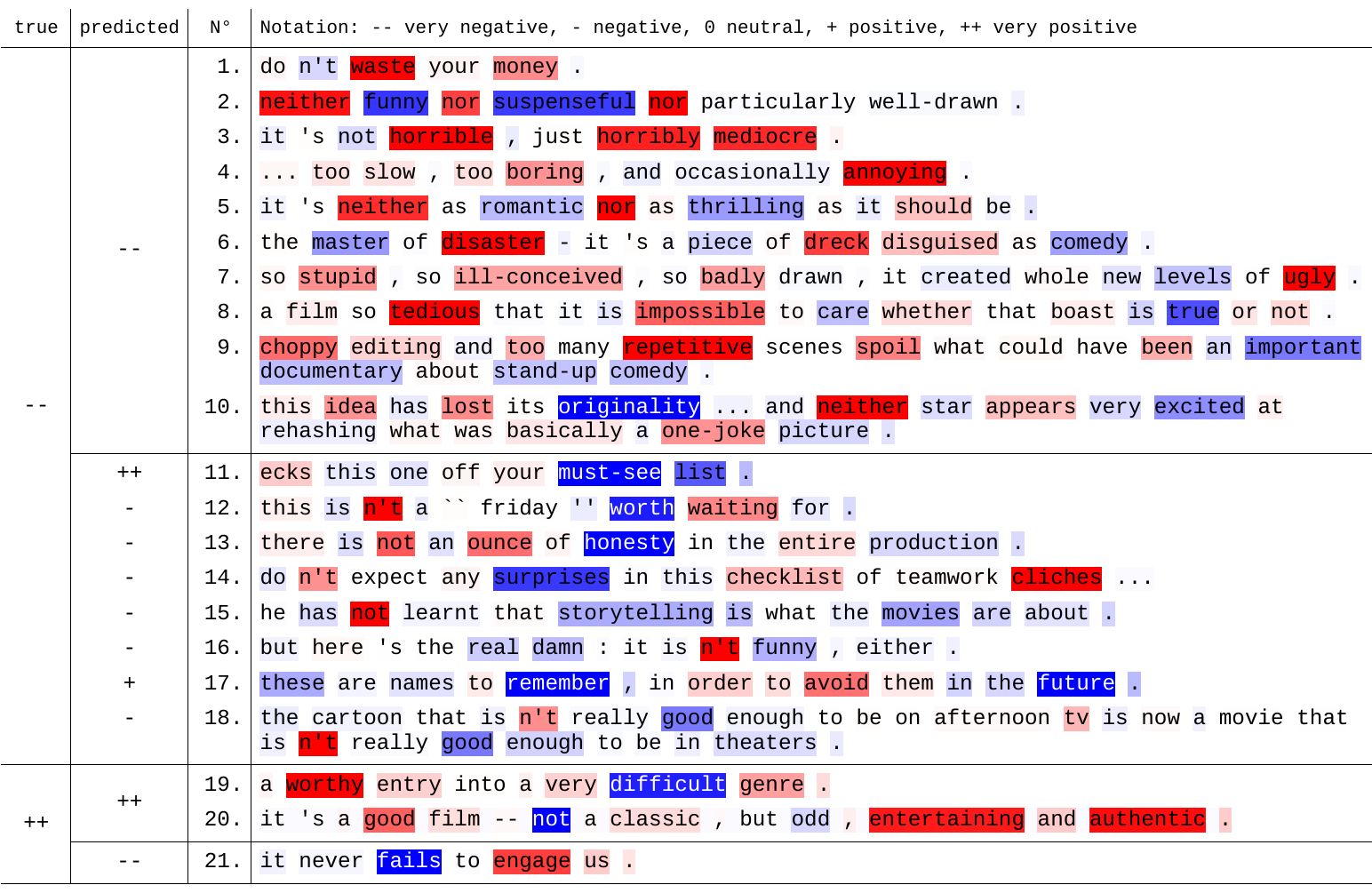}
\caption{ LRP heatmaps of exemplary test sentences, using as target class the {\it true} sentence class. Positive relevance is mapped to red, negative to blue, and the color intensity is normalized to the maximum absolute relevance per sentence. The true sentence class, and the classifier's predicted class, are indicated on the left.}\label{fig:heatmap_LRP}
\end{figure*}

In order to illustrate the differences between SA and LRP, we provide in Fig.~\ref{fig:heatmap_SA}~and~\ref{fig:heatmap_LRP} heatmaps of exemplary test set sentences. These heatmaps were obtained by mapping positive word-level relevance values to red, and negative relevances to blue.
The exemplary sentences belong either to the class ``very negative'' or to the class ``very positive'', and the target class for relevance decomposition is always the {\it true} class. On the left of the Figures, we indicate the {\it true} sentence class, as well as the bi-LSTM's {\it predicted} class, whereby the upper examples are correctly classified while the bottom examples are falsely classified.

From the inspection of the heatmaps, we notice that SA does not clearly distinguish between words speaking {\it for} or {\it against} the target class. Indeed it sometimes attributes a comparatively high relevance to words expressing a positive appreciation like {\it thrilling} (example 5), {\it master} (example  6) or {\it must-see} (example 11), while the target class is ``very negative''; or to the word {\it difficult} (example 19) expressing a negative judgment, while the target class is ``very positive''.
On the contrary, LRP can discern more reliably between words addressing a negative sentiment, such as {\it waste} (1), {\it horrible} (3), {\it disaster} (6), {\it repetitive} (9) (highlighted in red), or {\it difficult} (19) (highlighted in blue), from words indicating a positive opinion, like {\it funny} (2), {\it suspenseful} (2), {\it romantic} (5), {\it thrilling} (5) (highlighted in blue), or {\it worthy} (19), {\it entertaining} (20) (highlighted in red).

Furthermore, LRP explains well the two sentences that are mistakenly classified as ``very positive'' and ``positive'' (examples 11 and 17), by accentuating the negative relevance (blue colored) of terms speaking {\it against} the target class, i.e. the class ``very negative'', such as {\it must-see list}, {\it remember} and {\it future}, whereas such understanding is not provided by the SA heatmaps.
The same holds for the misclassified ``very positive'' sentence (example 21), where the word {\it fails} gets attributed a deep negatively signed relevance (blue colored).
A similar limitation of gradient-based relevance visualization for explaining predictions of recurrent models was also observed in previous work \cite{Li_2016a}.

Moreover, an interesting property we observe with LRP, is that the sentiment of negation is modulated by the sentiment of the subsequent words in the sentence. Hence, e.g. in the heatmaps for the target class ``very negative'', when negators like {\it n't} or {\it not} are followed by words indicating a negative sentiment like {\it waste}~(1) or {\it horrible}~(3), they are marked by a negatively signed relevance (blue colored), while when the subsequent words express a positive impression like {\it worth}~(12), {\it surprises}~(14), {\it funny}~(16) or {\it good}~(18), they get a positively signed relevance (red colored).

Thereby, the heatmap visualizations provide some insights on how the sentiment of single words is composed by the bi-LSTM model, and indicate that the sentiment attributed to words is not static, but depends on their context in the sentence.
Nevertheless, we would like to point out that the explanations delivered by relevance decomposition highly depend on the quality of the underlying classifier, and can only be ``as good'' as the neural network itself, hence a more carefully tuned model might deliver even better explanations.

\subsection{Representative Words for a Sentiment}

Another qualitative analysis we conduct is dataset-wide, and consists in building a list of the most resp. the least relevant words for a class. To this end, we first perform SA and LRP on {\it all} test set sentences for one specific target class, as an example we take the class ``very positive''.
Secondly, we order all words appearing in the test sentences in decreasing resp. in increasing order of their relevance value, and retrieve in Table~\ref{fig:keywords} the ten most and least relevant words we obtain.
From the SA word lists, we observe that the highest SA relevances mainly point to words with a strong semantic meaning, but not necessarily expressing a positive sentiment, see e.g. {\it broken-down}, {\it lackadaisical} and {\it mournfully}, while the lowest SA relevances correspond to stop words.
On the contrary, the extremal LRP relevances are more reliable: the highest relevances indicate words expressing a positive sentiment, while the lowest relevances are attributed to words defining a negative sentiment, hence both extremal relevances are related in a meaningful way to the target class of interest, i.e. the class ``very positive''.

\begin{table}
\resizebox{\columnwidth}{!}{%
\begin{tabular}{ |l|l|l|l| }
  \hline
  \multicolumn{2}{|c|}{SA} & \multicolumn{2}{|c|}{LRP} \\
  \hline
  most relevant 	& least relevant 	& most relevant 	& least relevant \\
  \hline
  broken-down 	        & into 			& funnier 		& wrong 	 \\
  wall 	                & what 			& charm 	        & n't 	         \\
  execution 	        & that 			& polished	        & forgettable 	 \\
  lackadaisical 	& a 			& gorgeous	        & shame 	 \\
  milestone      	& do 			& excellent	        & little 	 \\
  unreality      	& of 			& screen 	        & predictable	 \\
  soldier       	& all 			& honest 	        & overblown	 \\
  mournfully    	& ca 			& wall 	        	& trying	 \\
  insight       	& in 			& confidence 	        & lacking	 \\
  disorienting  	& 's 			& perfectly	        & nonsense 	 \\
  \hline
  \end{tabular}}
\caption{Ten most resp. least relevant words identified by SA and LRP over all 2210 test sentences, using as relevance target class the class ``very positive''.} \label{fig:keywords}
\end{table}

\subsection{Validation of Word Relevance}

\begin{figure}
\centering
\includegraphics[width=1.0\columnwidth]{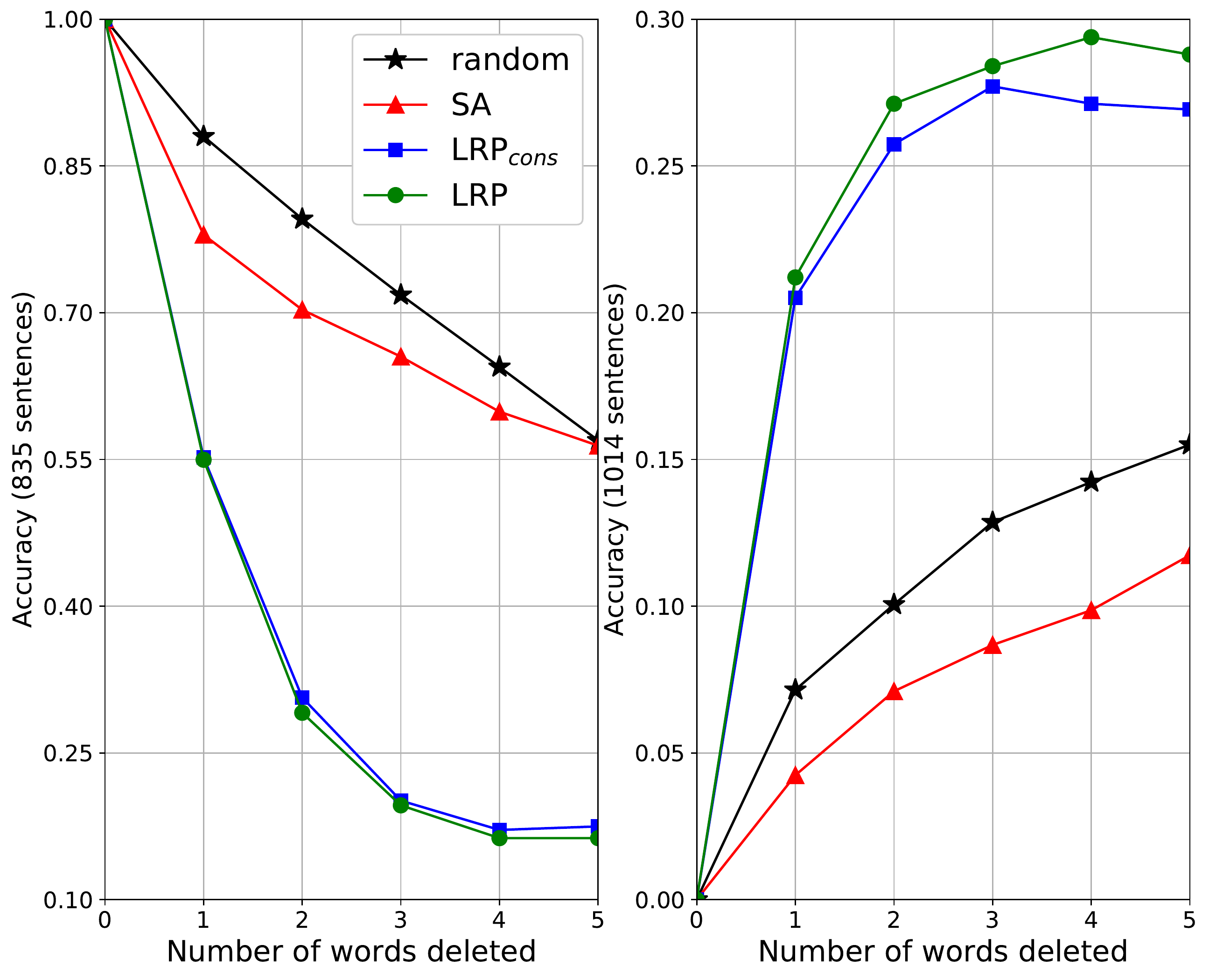}
\caption{Impact of word deleting on initially correctly (left) and falsely (right) classified test sentences, using either SA or LRP as relevance decomposition method (LRP$_{cons}$ is a variant of LRP with relevance conservation). The relevance target class is the true sentence class, and words are deleted in decreasing (left) and increasing (right) order of their relevance. Random deletion is averaged over 10 runs (std $<$ 0.016). A steep decline (left) and incline (right) indicate informative word relevance.}\label{fig:deletion}
\end{figure}

In order to quantitatively validate the word-level relevances obtained with SA and LRP, we perform two word deleting experiments.
For these experiments we consider only test set sentences with a length greater or equal to 10 words (this amounts to retain 1849 test sentences), and we delete from each sentence up to 5 words accordingly to their SA resp. LRP relevance value (for deleting a word we simply set its word embedding to zero in the input sentence representation), and re-predict via the bi-LSTM the sentiment of the sentence with ``missing'' words, to track the impact of these deletions on the classifier's decision.
The idea behind this experiment is that the relevance decomposition method that most pertinently reveals words that are important to the classifier's decision, will impact the most this decision when deleting words accordingly to their relevance value.
Prior to the deletions, we first compute the SA resp. LRP word-level relevances on the original sentences (with no word deleted), using the {\it true} sentence sentiment as target class for the relevance decomposition.
Then, we conduct two types of deletions. On initially correctly classified sentences we delete words in decreasing order of their relevance value, and on initially falsely classified sentences we delete words in increasing order of their relevance.
We additionally perform a random word deletion as an uninformative variant for comparison.
Our results in terms of tracking the classification accuracy over the number of word deletions per sentence are reported in Fig.~\ref{fig:deletion}.
These results show that, in both considered cases, deleting words in decreasing or increasing order of their LRP relevance has the most pertinent effect, suggesting that this relevance decomposition method is the most appropriate for detecting words speaking {\it for} or {\it against} a classifier's decision.
While the LRP variant with relevance conservation LRP$_{cons}$ performs almost as good as standard LRP, the latter yields slightly superior results and thus should be preferred. Finally, when deleting words in increasing order of their relevance value starting with initially falsely classified sentences (Fig.~\ref{fig:deletion} right), we observe that SA performs even worse than random deletion. This indicates that the lowest SA relevances point essentially to words that have no influence on the classifier's decision at all, rather that signalizing words that are ``inhibiting'' it's decision and speaking {\it against} the true class, as LRP is indeed able to identify.
Similar conclusions were drawn when comparing SA and LRP on a convolutional network for document classification \cite{arras-acl16}.

\subsection{Relevance Distribution over Sentence Length}

\begin{figure*}[th]
\centering
\includegraphics[width=1.0\textwidth]{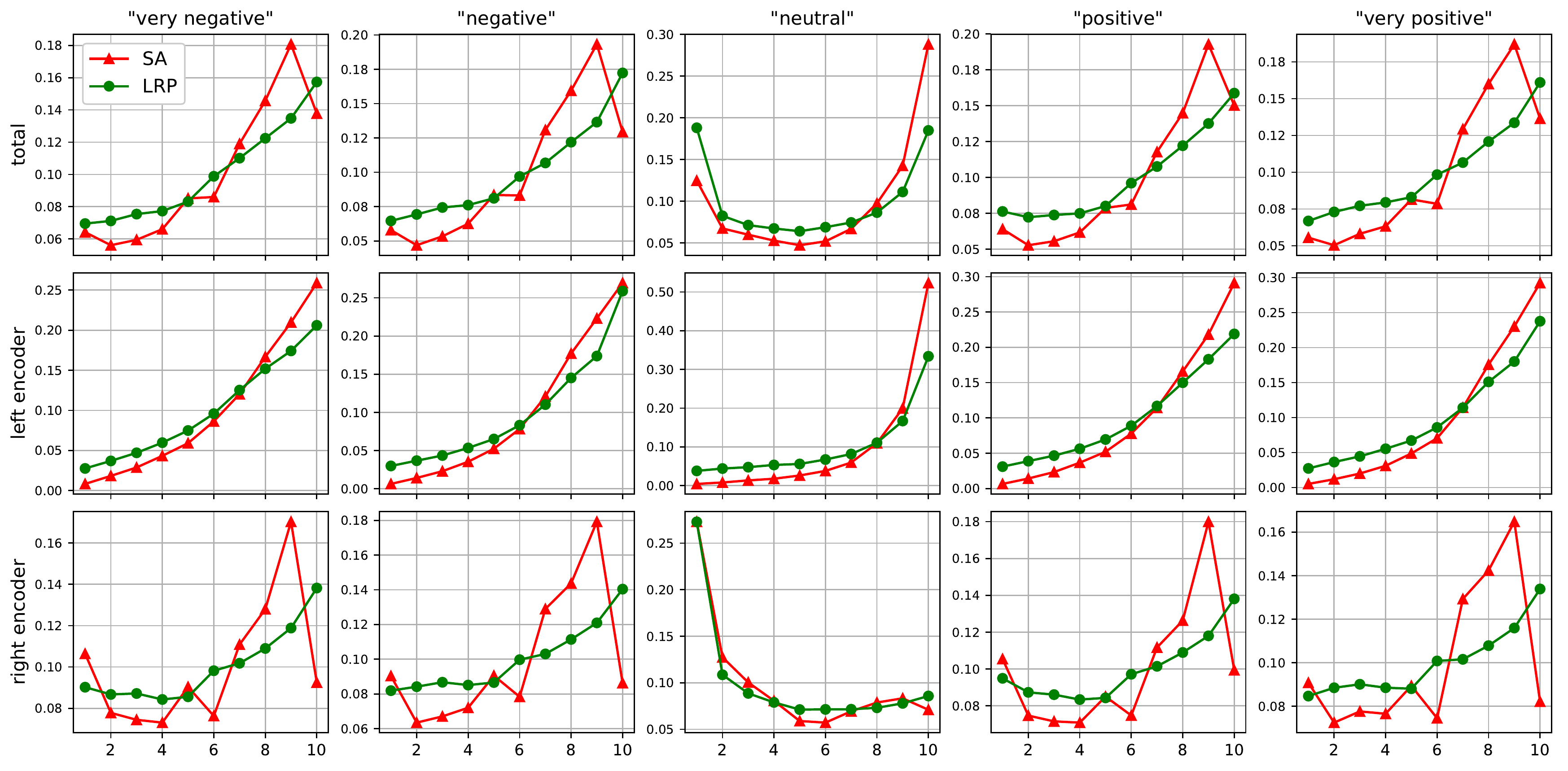}
\caption{Word relevance distribution over the sentence length (divided into 10 intervals), per relevance target class (indicated on the top), obtained by performing SA and LRP on all test sentences having a length greater or equal to 19 words (1104 sentences). For LRP, the absolute value of the word-level relevances is used to compute these statistics. The first row corresponds to the total relevance, the second resp. third row only contain the relevance from the bi-LSTM's left and right encoder.}\label{fig:relevance_distr}
\end{figure*}

To get an idea of which words over the sentence length get attributed the most relevance, we compute a word relevance statistic by performing SA and LRP on all test sentences having a length greater or equal to 19 words (this amounts to 50.0\% of the test set).
Then, we divide each sentence length into 10 equal intervals, and sum up the word relevances in each interval (when a word is not entirely in an interval, the relevance portion falling into that interval is considered). For LRP, we use the absolute value of the word-level relevance values (to avoid that negative relevances cancel out positive relevances). Finally, to get a distribution, we normalize the results to sum up to one.
We compute this statistic by considering either the total word relevance obtained via the bi-LSTM model, or by considering only the part of the relevance that comes from one of the two unidirectional model constituents, i.e. the relevance contributed by the LSTM which takes as input the sentence words in their original order (we call it left encoder), or the one contributed by the LSTM which takes as input the sentence words in reversed order (we call it right encoder).
The resulting distributions, for different relevance target classes, are reported in Fig.~\ref{fig:relevance_distr}.
Interestingly, the relevance distributions are not symmetric w.r.t. to the sentence middle, and the major part of the relevance is attributed to the second half of the sentences, except for the target class ``neutral'', where the most relevance is attributed to the last computational time steps of the left or the right encoder, resulting in an almost symmetric distribution of the total relevance for that class.
This can maybe be explained by the fact that, at least for longer movie reviews, strong judgments on the movie's quality tend to appear at the end of the sentences, while the beginning of the sentences serves as an introduction to the review's topic, describing e.g. the movie's subject or genre.
Another particularity of the relevance distribution we notice, is that the relevances of the left encoder tend to be more smooth than those of the right encoder, which is a surprising result, as one might expect that both unidirectional model constituents behave similarly, and that there is no mechanism in the model to make a distinction between the text read in original and in reversed order.


\section{Conclusion}
In this work we have introduced a simple yet effective strategy for extending the LRP procedure to recurrent architectures, such as LSTMs, by proposing a rule to backpropagate the relevance through multiplicative interactions.
We applied the extended LRP version to a bi-directional LSTM model for the sentiment prediction of sentences, demonstrating that the resulting word relevances trustworthy reveal words supporting the classifier's decision {\it for}
or {\it against} a specific class, and perform better than those obtained by a gradient-based decomposition.

Our technique helps understanding and verifying the correct behavior of recurrent classifiers, and can detect important patterns in text datasets.
Compared to other non-gradient based explanation methods, which rely e.g. on random sampling or on iterative representation occlusion, our technique is deterministic, and can be computed in one pass through the network. Moreover, our method is self-contained, in that it does not require to train an external classifier to deliver the explanations, these are obtained directly via the original classifier.

Future work would include applying the proposed technique to other recurrent architectures such as character-level models or GRUs, as well as to extractive summarization.
Besides, our method is not restricted to the NLP domain, and might also be useful to other applications relying on recurrent architectures.

\section*{Acknowledgments}
We thank Rico Raber for many insightful discussions. This work was partly supported by BMBF, DFG and also Institute for Information \& Communications Technology Promotion (IITP) grant funded by the Korea government (No.\ 2017-0-00451 for KRM).

\bibliography{bibliography}
\bibliographystyle{emnlp_natbib}

\section*{Appendix}

\noindent{\bf Long-Short Term Memory Network (LSTM)}
We define in the following the LSTM recurrence equations \cite{Hochreiter_1997, Gers_2000} of the model we used in our experiments:
\begin{equation*}
\begin{split}
  i_t &= \texttt{sigm} \;\; \Big( W_i \; h_{t-1} + U_i \; x_t + b_i \Big) \\
  f_t &= \texttt{sigm} \;   \Big( W_f \; h_{t-1} + U_f \; x_t + b_f \Big) \\
  o_t &= \texttt{sigm} \;   \Big( W_o \; h_{t-1} + U_o \; x_t + b_o \Big) \\
  g_t &= \texttt{tanh} \;   \Big( W_g \; h_{t-1} + U_g \; x_t + b_g \Big) \\
  c_t &= f_t \odot c_{t-1} \;  + \; i_t \odot g_t     \\
  h_t &= o_t \odot  \texttt{tanh} (c_t)
\end{split}
\end{equation*}
Here above the activation functions $\texttt{sigm}$ and $\texttt{tanh}$ are applied element-wise, and $\odot$ is an element-wise multiplication.

As an input, the LSTM gets fed with a sequence of vectors $\boldsymbol{x} = (x_1, x_2,..., x_T)$ representing the word embeddings of the input sentence's words. The matrices $W$'s, $U$'s, and vectors $b$'s are connection weights and biases,
and the initial states $h_0$ and $c_0$ are set to zero. 

The last hidden state $h_T$ is eventually attached to a fully-connected linear layer yielding a prediction score vector $\boldsymbol{f}(\boldsymbol{x})$, with one entry ${f_c}(\boldsymbol{x})$ per class, which is used for sentiment prediction.

\noindent{\bf Bi-directional LSTM}
The bi-directional LSTM \cite{Schuster_1997} we use in the present work, is a concatenation of two separate LSTM models as described above, each of them taking a different sequence of word embeddings as input. 

One LSTM takes as input the words in their original order, as they appear in the input sentence.
The second LSTM takes as input the same words but in {\it reversed} order.

Each of these LSTMs yields a final hidden state vector, say  $h^{\rightarrow}_T$ and $h^{\leftarrow}_T$. The concatenation of these two vectors is eventually fed to a fully-connected linear layer, retrieving one prediction score ${f_c}(\boldsymbol{x})$ per class.

\end{document}